\documentclass[runningheads]{llncs}
\usepackage{graphicx}
\usepackage{comment}
\usepackage{amsmath,amssymb} 
\usepackage{color}

\usepackage{soul}
\usepackage{url}
\usepackage[hidelinks]{hyperref}

\usepackage{amsmath}
\usepackage{amsfonts}
\usepackage{booktabs}
\usepackage{algorithm}
\usepackage{algorithmic}
\usepackage{color}
\usepackage{multirow}
\usepackage{subfigure}
\usepackage{hyperref}
\newcommand{\tabincell}[2]{\begin{tabular}{@{}#1@{}}#2\end{tabular}}

\begin{document}
\pagestyle{headings}
\mainmatter
\def\ECCVSubNumber{6404}  

\newcommand{\customcomment}[3]{\textcolor{#1}{[#2:#3]}}

\newcommand{\todo}[1]{{\color{red}#1}}
\newcommand{\chj}[1]{{\color{black}#1}}
\newcommand{\cjy}[1]{{\color{black}#1}}

\title{Generative Adversarial Zero-shot Learning via Knowledge Graphs} 

\titlerunning{Generative Adversarial Zero-shot Learning via Knowledge Graphs}
%
\author{
Yuxia Geng\inst{1}
\and
Jiaoyan Chen\inst{2}
\and
Zhuo Chen\inst{1}
\and
Zhiquan Ye\inst{1}
\and
Zonggang Yuan\inst{3}
\and
Yantao Jia\inst{3}
\and
Huajun Chen\inst{1}
}
\authorrunning{Y. Geng et al.}
%
\institute{College of Computer Science and Technology, Zhejiang University, Hangzhou, China\\
\email{\{gengyx,chenzhuo98,yezq,huajunsir\}@zju.edu.cn}
\and
Department of Computer Science, University of Oxford, Oxford, UK
\email{jiaoyan.chen@cs.ox.ac.uk}
\and
Huawei Technologies Co., Ltd, China
\\
\email{yuanzonggang@huawei.com,jamaths.h@163.com}
}
\maketitle

\begin{abstract}

Zero-shot learning (ZSL) is to handle the prediction of those unseen classes that have no labeled training data.
	Recently, generative methods like Generative Adversarial Networks (GANs) are being widely investigated for ZSL due to their high accuracy, generalization capability and so on.
	However, the side information of classes used now is limited to text descriptions and attribute annotations, which are in short of semantics of the classes.
	In this paper, we introduce a new generative ZSL method named KG-GAN by incorporating rich semantics in a knowledge graph (KG) into GANs.
	Specifically, we build upon Graph Neural Networks and encode KG from two views: {\it class} view and {\it attribute} view considering the different semantics of KG.
	With well-learned semantic embeddings for each node (representing a visual category), we leverage GANs to synthesize compelling visual features for unseen classes.
	According to our evaluation with multiple image classification datasets, KG-GAN can achieve better performance than the state-of-the-art baselines.

\end{abstract}

\section{Introduction}

\begin{figure}
\centering
\includegraphics[height=4cm]{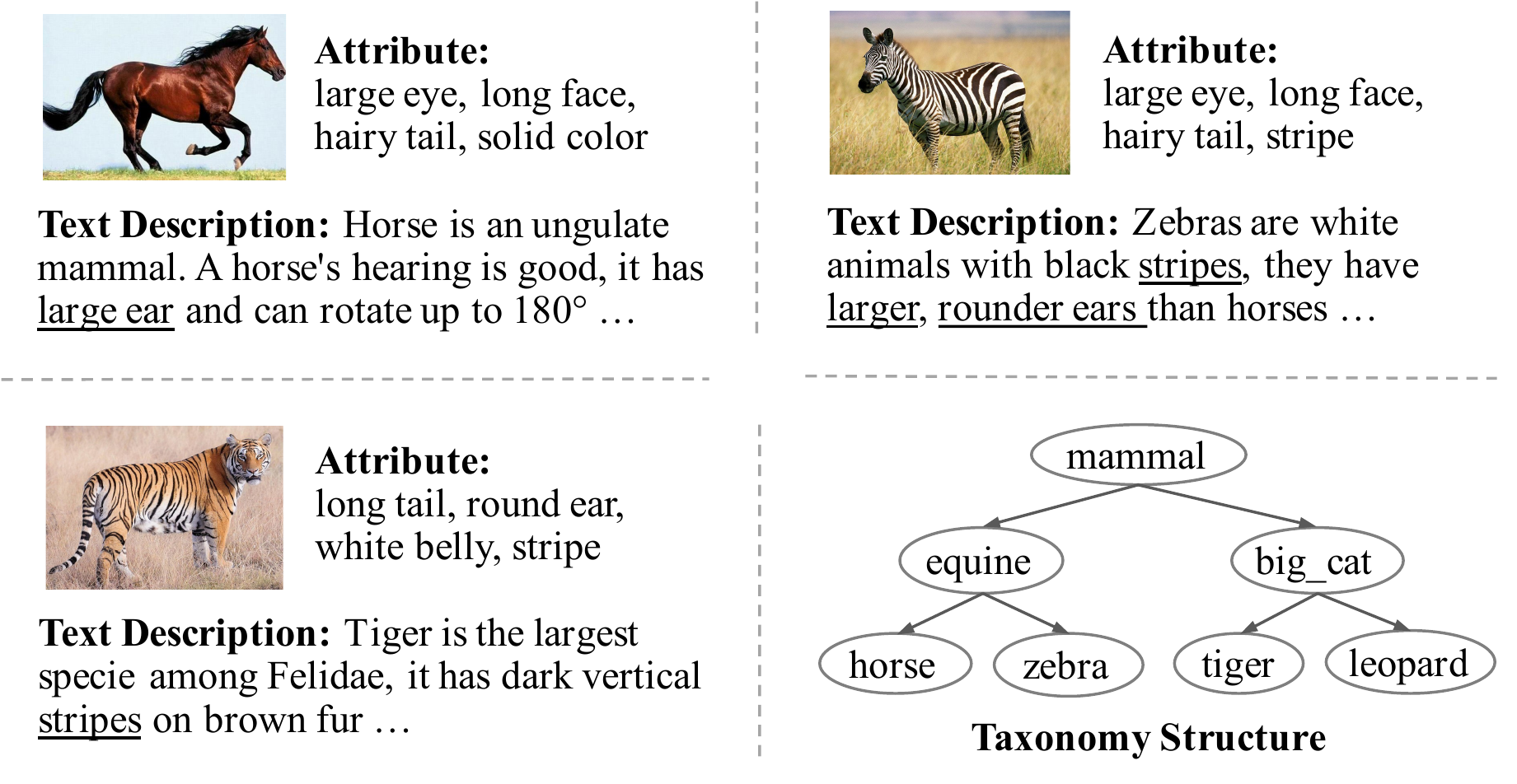}
\caption{The general side information of three classes: {\it horse}, {\it zebra} and {\it tiger}, including attribute annotations, text descriptions and their relationship in taxonomy structure.}
\label{fig:example}
\end{figure}

Machine learning often operates on a closed world assumption: it trains the model with a number of labeled samples and makes prediction with classes that have appeared in the training stage (i.e., seen classes) alone.
This limitation raises a hot research interest in Zero-shot Learning (ZSL), which aims to handle novel classes without any training samples (i.e., unseen classes). 
An intuitive idea to deal with such unseen classes is to take advantages of {\it side information} of classes, which builds the semantic relationships among classes and enables transferring knowledge obtained from seen classes to the unseen. For example, in classifying animal images, the side information usually contains the visual characteristics of classes (e.g., human-annotated attributes or textual descriptions from external sources like Wikipedia) and how unseen classes are related to seen classes (e.g., class hierarchy in taxonomy), as Fig.~\ref{fig:example} shows.

ZSL is often divided into two paradigms. One is based on mapping \cite{lampert2009learning,norouzi2013zero,DBLP:conf/cvpr/KodirovXG17,DBLP:conf/cvpr/0004YG18}.
It learns a general mapping function to map visual features and/or semantic features into the same latent space and then conduct nearest neighbor search to predict the class labels. However, it is a non-trivial task to bridge the {\it semantic gap} between such two spaces since the class-level semantic descriptions produce non-visual features. Additionally, the nearest neighbor search suffers from the {\it hubness} problem, that is, the neighborhoods of the mapped elements are biased to some hubs vectors, pushing the correct labels down the neighbor list \cite{DBLP:journals/jmlr/RadovanovicNI10}.

The other ZSL paradigm is developed upon generative models such as generative adversarial network (GANs) \cite{DBLP:conf/cvpr/XianLSA18,DBLP:conf/cvpr/ZhuEL0E18,DBLP:conf/cvpr/HuangWYW19,DBLP:conf/cvpr/LiJLD0H19,zhu2019learning,sariyildiz2019gradient}.
These methods utilize side information of classes to synthesize samples (or features) for unseen classes, circumvent the need for space mapping, thus avoiding the {\it semantic gap} problem as well. 
Such a generative solution transforms the ZSL problem into a traditional supervised learning problem which can be flexibly dealt with by kinds of existing methods.
Moreover, the generated unseen samples (or features) can alleviate the training bias towards seen classes and avoid the above mentioned {\it hubness} problem.

However, most of these generative methods are built upon one type of side information such as attribute annotations, taxonomy structure or textual descriptions.
Thus, they often generate less discriminative samples (or features) without enough variation.
Considering the example of human-annotated attributes of animal classes,
when we use attribute ``stripe" to generate samples for class {\it zebra}, another significantly different class {\it tiger} which is also annotated with ``stripe" may also obtain synthesized samples with similar features as {\it zebra} (i.e., domain-shift problem \cite{DBLP:journals/pami/FuHXG15}), 
especially when {\it tiger} lacks of other representative attribute annotations.
Taxonomy structure describes the inter-class relationship in taxonomy, e.g., {\it horse} belongs to {\it equine} while {\it tiger} belongs to {\it big cat}.
However, it will generate less discriminative samples for sibling classes  which may look quite different, such as {\it horse} and {\it zebra}.
Textual descriptions can be easily collected but are prone to introduce much noise due to the ambiguity and irrelevant words and phrases.

\cjy{In this paper, we propose to incorporate a knowledge graph (KG) which contains semantics of all the above mentioned side information and can lead to a higher ZSL performance.}
To this end, we first build the KG with two views\footnote{Note that a KG can be constructed by various automatic and semi-automatic tools with domain knowledge and domain experts.}: a {\it class} view for taxonomy structure and an {\it attribute} view for attribute annotations, 
and embed it into a vector space with Graph Neural Networks (GNNs) together with the class name word vector learned from textual descriptions. 
We then propose KG-GAN -- a new generative ZSL framework utilizing the above KG embeddings, and Generative Adversarial Networks (GANs) which synthesize discriminative visual features for each class.
Unlike previous generative methods using engineered regularizers or complex networks, our framework adopts the basic GAN model without any additions.
Our main contributions are as follows:
\begin{itemize}
	\item As far as we know, KG-GAN is among the first to utilize formal and rich semantics represented by a KG in generative ZSL. The KG describes different aspects of ZSL classes, promoting the knowledge transfer between seen and unseen classes.
	\item We develop Graph Neural Networks to learn semantically meaningful class embeddings so as to investigate how class semantics influences the feature transfer in ZSL.
	\item 
	\cjy{We contribute two new ZSL benchmarks on image classification as well as their corresponding KGs.
	Experiments on these two benchmarks show that the generated features are quite effective to both seen and unseen classes, and promising results have been achieved in comparison with the state-of-the-art baselines including both generative and none generative ZSL methods.}
\end{itemize}

\section{Related Work}

\subsection{Mapping-based vs Generative ZSL}
In mapping-based zero-shot learning literature, methods like \cite{frome2013devise,lampert2013attribute,norouzi2013zero,DBLP:conf/cvpr/KodirovXG17} tried to map visual features into semantic space, and found the most similar class by computing nearest neighbor on class embeddings.
However, these methods tend to aggravate {\it hubness} problem \cite{DBLP:conf/pkdd/ShigetoSHSM15} because a number of visual features are mapped into a point in the semantic space for a certain class, leading to an increasing probability of irrelevant points (hubs) being the nearest neighbors (i.e., the correct labels).
Some other methods proposed to map class embeddings into visual space to suppress this problem \cite{DBLP:journals/corr/DinuB14,DBLP:conf/cvpr/ZhangXG17,DBLP:conf/cvpr/ChangpinyoCGS16}, in which the features of unseen classes are learned by transferring features of seen classes based on the class relatedness in semantic space. 
However, the feature transfer is restricted by the mapping of two spaces and heavily dependent on the semantic relationships of classes, which may be undermined by the {\it semantic gap} problem.
Besides, the learning of unseen classifiers only rely on the samples of seen classes, which may have strong bias towards seen classes during prediction.

Different from mapping-based ZSL, generative zero-shot learning directly synthesizes unseen samples (or features) with random noises which are conditioned by the class side information.
For example, Zhu et al.
\cite{DBLP:conf/cvpr/ZhuEL0E18} utilized GANs as generative models which took the textual descriptions of classes from Wikipedia articles as input and generated visual features for these classes,
with a fully connected layer being used to reduce text noise.
Additionally, they also proposed a visual pivot regularization to preserve the inter-class discrimination of generated features. 
Huang et al. \cite{DBLP:conf/cvpr/HuangWYW19} introduced a generator to synthesize sample features with class embeddings and a regressor to project generated features back to their corresponding class embeddings, while a discriminator was to evaluate the closeness of visual features and class embeddings.
However, most of these methods rely on engineered regularizers or auxiliary networks to guarantee the quality of generated samples.
Few of them consider the effectiveness of class semantics.
In our paper, we leverage knowledge graph which contains rich class semantics to enhance the feature generation in ZSL, making the synthesized data discriminative and variational. 

\subsection{KG-based ZSL}
There have been some studies that utilize KGs to enhance mapping-based ZSL.
For instance, Wang et al. \cite{DBLP:conf/cvpr/0004YG18} proposed a KG which depicted the class hierarchy in taxonomy, and used this hierarchical relationship to predict classifiers for unseen classes, where Graph Convolutional Network (GCN) was applied to transfer features from seen classifiers to unseen classifiers.
This method as well as its derivative \cite{DBLP:conf/cvpr/KampffmeyerCLWZ19} have the following problems:
1) the KG used is homogeneous, where the class semantics is limited especially for discriminating those sibling classes;
2) the models are trained only using seen samples and have strong bias towards seen classes during prediction, especially in the generalized ZSL scenario where the prediction involves both seen and unseen class labels.
In contrast, we use a KG with richer semantics from various side information including class structure in taxonomy, class attributes and class name word vectors,
while the generative solution which generates training samples for unseen classes is free of the bias issue in the generalized ZSL scenario.
As far as we know, there are currently no works that incorporate such a KG in generative ZSL.

\section{Methodology}

\begin{figure}
\centering
\includegraphics[height=4cm]{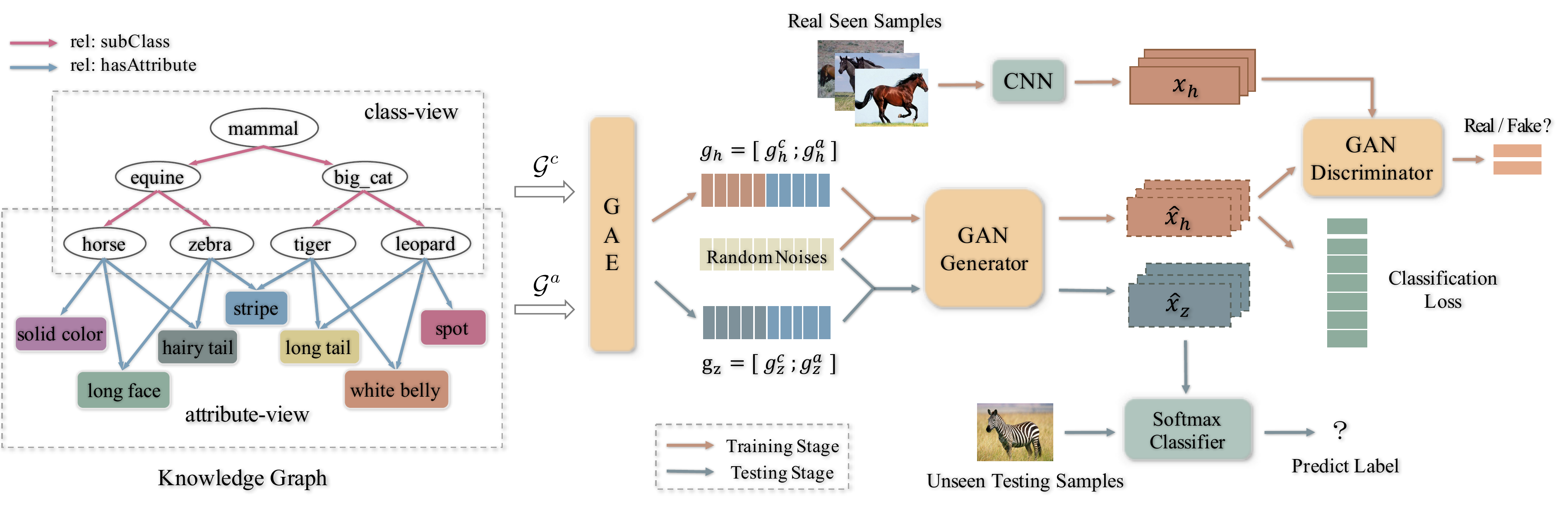}
\caption{An overview of our KG-GAN framework. 
    The left is the knowledge graph consisting of nodes (i.e., {\it classes} and {\it attributes}) and relation edges (i.e., {\it subClass} and {\it hasAttribute}).
    The right is the GAN module.
    $g_h$ and $g_z$ represent the class embeddings of {\it horse} and {\it zebra} respectively, which consist of embeddings learned from {\it class}-view graph $\mathcal{G}^c$ and {\it attribute}-view graph $\mathcal{G}^a$.
    {\it horse} is seen in the training set, $x_h$ and $\hat{x}_h$ are its real image features and synthesized image features, while {\it zebra} is unseen, whose features $\hat{x}_z$ are synthesized to learn classifier for it at testing stage.}
\label{fig:framework}
\end{figure}

In this section, we introduce KG-GAN in details, as shown in Fig.~\ref{fig:framework}.
We first utilize unsupervised graph neural networks, i.e., Graph Auto-Encoders (GAEs), to embed our knowledge graph which includes two views: the {\it class} view which models the hierarchical relationship among classes, and the {\it attribute} view which models the human annotated visual characteristics.
Briefly, we first obtain one embedding vector for each class according to its corresponding node in the KG.
Secondly, a GAN with Wasserstein distance \cite{arjovsky2017wasserstein} and a classification loss is adopted for feature generation.
It includes (i) a generator for synthesizing visual features from random noises that are conditioned on the class embedding,
(ii) a discriminator for distinguishing the generated features from real ones,
and (iii) a supervised classification loss for discriminating classes with the generated features. 
Note that we generate image features instead of raw images for both higher accuracy and computation efficiency \cite{DBLP:conf/cvpr/XianLSA18},
and we adopt some pre-trained CNN models (e.g., ResNet) for extracting image features.
Finally, with well-trained generator, we can synthesize compelling image features for each unseen class and train a softmax classifier for it.

\subsection{Preliminaries: ZSL and KG}

We first formalize the ZSL problem as follows.
Let $\mathcal{D}_{tr} = \{(x, y) | x \in \mathcal{X}_s, y \in \mathcal{Y}_s\}$ be the training set of ZSL, 
where $x$ is the CNN feature of training image,
$y$ represents the class label in $\mathcal{Y}_s$ consisting of seen classes.
While the testing set is denoted as $\mathcal{D}_{te} = \{(x, y) | x \in \mathcal{X}_u, y \in \mathcal{Y}_u\}$,
where $\mathcal{Y}_u$, the set of unseen classes, has no overlap with $\mathcal{Y}_{s}$.
At training stage, 
$\mathcal{D}_{tr}$ and $\mathcal{Y}_u$ are usually used to learn one classifier for each unseen class.
We study two settings in prediction: the ZSL setting and the generalized ZSL (GZSL) setting.
The former is to predict the label of testing samples in $\mathcal{X}_u$ with candidates from $\mathcal{Y}_u$, 
while the latter extends the testing set to $\mathcal{X}_s \cup \mathcal{X}_u$, with candidate labels from both seen classes and unseen classes i.e., $\mathcal{Y}_s \cup \mathcal{Y}_u$.

A KG is used as additional input for the above mentioned training.
The KG, denoted as $\mathcal{G}$, includes
a set of class nodes, denoted as $\mathcal{C} =\{c_1, c_2, ..., c_n\}$, and a set of attribute nodes, denoted as $\mathcal{A} =\{a_1, a_2, ..., a_m\}$, as the left of Fig.~\ref{fig:framework} shows.
The KG is also composed of two views (parts): {\it class}-view denoted as $\mathcal{G}^c$, and {\it attribute}-view denoted as $\mathcal{G}^a$.
$\mathcal{G}^c$ is formed by the class nodes in $\mathcal{C}$, where a relation (edge) named ``{\it subClass}" is used to model the hierarchical relationship of classes defined in taxonomy.
$\mathcal{G}^a$ is a heterogeneous bipartite graph consisting of class nodes and attribute nodes, where each class node is connected with a series of attributes nodes annotated for it via ``{\it hasAttribute}" relation edge.
In KG embedding, we learn a function $g(\cdot)$ to encode each class node as a vector known as {\it class embedding}.
For each class $y$ in $\mathcal{Y}_s \cup \mathcal{Y}_u$, we can learn two embeddings $g^c(y)$ and $g^a(y)$, according to the {\it class}-view and the {\it attribute}-view respectively.
For convenience, the embeddings of $i$-th class are denoted with the subscript, i.e., $g^c_i$ and $g^a_i$.

\subsection{Class Embedding from KG}\label{class_embed}
Graph Auto-Encoder (GAE) \cite{kipf2016variational} is a method for unsupervised learning on graph-structured data based on the variational auto-encoder (VAE) \cite{DBLP:journals/corr/KingmaW13}.
It takes graph convolutional network (GCN) as encoder and inner product as decoder, enabling the latent representations of graph nodes to be learned, which are the class embeddings we desire.

\subsubsection{Graph Encoder}
GCN works on propagating information between the nodes in graph via a series of graph convolutions and capturing the dependence of graph-structured data.
We therefore use GCN to encode the inter-class relationship and the class-attribute correlation reflected in the proposed KG.

In each layer of GCN, the convolutional operation computes a node's vector representation by aggregating the vectors of its neighboring nodes defined in the graph, and update it to the next layer.
Stacking the convolutional operation one after another, we can output the latent embeddings of graph nodes at last layer.

Given the KG $\mathcal{G}$, we first apply GCN on graph $\mathcal{G}^c$ and compute the embeddings of class nodes from the {\it class}-view.
For class $i$, its $k$-th layer vector is represented as: 
\begin{equation}
	g^c_{i,k} = \sigma (W_k^c \sum_{j \in N_i} \frac{g^c_{j,k-1}}{|N_i|} + B_k^c g^c_{i,k-1}) 
\end{equation}
where $N_i$ is the set of neighboring classes of class $i$, $W_k^c$ and $B_k^c$ denote the layer-specific trainable weight matrix and bias term in {\it class}-view, respectively.
The latent embedding of class $i$ from class-view is outputted at last layer: 
$g_i^c = g^c_{i,K}$.

We then apply another GCN on graph $\mathcal{G}^a$, where the vectors of class nodes are aggregated by the vectors of their neighboring attribute nodes.
Similar with the {\it class}-view, the vector of class $i$ in the $l$-th layer is computed as:
\begin{equation}
	g^a_{i,l} = \sigma (W_l^a \sum_{j \in M_i} \frac{g^a_{j,l-1}}{|M_i|} + B_k^a g^a_{i,l-1}) 
\end{equation}
where $M_i$ is the set of neighboring attributes of class $i$ (i.e., the annotated attributes of class $i$), and $g^a_{j,l-1}$ is the vector representation of attribute $j$ at $(l-1)$-th layer. $W_l^a$ and $B_l^a$ represent the layer-specific weight and bias in {\it attribute}-view.
We also obtain the {\it attribute}-view latent embedding of class $i$ from the output of last layer of GCN: $g_i^a = g^a_{i,L}$.

Finally, we concatenate the {\it class}-view embedding $g_i^c$ and the {\it attribute}-view embedding $g_i^a$ to form the {\it class embedding} for class $i$:
\begin{equation}
	g_i = [g_i^c; g_i^a] 
\end{equation}

To enrich the semantics of graph nodes,
we initialize the node representations with word embeddings of class name and attribute name that are trained on skip-gram model on Wikipedia articles.
These embeddings are taken as the input of the two GCNs.

\subsubsection{Graph Decoder}
To preserve the relationship between two nodes that are connected via a relation edge, the decoder performs proximity calculation between these linked nodes in the latent space.
Specifically, for each linked node pair, we conduct the inner product between their latent embeddings.
With observed links in the graph, we can optimize the model by minimizing the following loss function:
\begin{equation}	
\mathcal{L} = - \sum_{(i,j)\in \Omega} log \ \sigma(g_i^\top \cdot g_j) - w \sum_{(i,j')\in \Omega^-} log \  \sigma(-g_{i}^\top \cdot g_{j'})
\end{equation}
where $\sigma(\cdot)$ is the sigmoid function, ($i$, $j$) is a pair of linked nodes, and $j'$ is a node not connected with $i$.
$\Omega$ is the observed (positive) link set, $\Omega^-$ is the negative set,
which involves all unlinked node pairs (the complement of $\Omega$), 
and $w$ is a weight computed by the ratio of number of positive and negative links.
In this way, we keep nodes with links to be close to each other and nodes without links far apart, and optimize the learning of latent representation of graph nodes.

\subsection{Feature Generation with GAN}
With well-learned semantically meaningful {\it class embeddings}, we learn a generator $G$, which takes class embedding $g(y)$ and random noise vector $z$ sampled from Gaussian distribution $\mathcal{N}(0,1)$ as its inputs and synthesizes a CNN image feature $\hat{x}$ of class $y$.
The loss of $G$ is defined as:
\begin{equation}
	\mathcal{L}_G = - \mathbb{E}[D(\hat{x})] - \lambda \mathbb{E}[logP(y|\hat{x})]
\end{equation}
where $\hat{x} = G(z, g(y))$. 
The first term of loss function is the Wasserstein loss \cite{arjovsky2017wasserstein}, and the second term is the supervised classification loss for classifying the synthesized features.
$\lambda$ is the corresponding weight coefficient.

The discriminator $D$ then takes synthesized features $\hat{x}$ and real features $x$ extracted from a training image of $y$ as input, the loss can be formulated as:
\begin{equation}
	\mathcal{L}_D = \mathbb{E}[D(x, g(y))] - \mathbb{E}[D(\hat{x})] - \beta \mathbb{E}[(|| \bigtriangledown_{\tilde{x}} D(\tilde{x}) ||_p -1)^2]
\end{equation}
where the first two terms approximate the Wasserstein distance of the distribution of real features and synthesized features, and the last term is the gradient penalty to enforce the gradient of $D$ to have unit norm (i.e., Lipschitz constraint proposed in \cite{DBLP:conf/nips/GulrajaniAADC17}), in which $\tilde{x} = \varepsilon x + (1-\varepsilon) \hat{x}$ with $\varepsilon \sim U(0,1)$, and $\beta$ is the corresponding weight coefficient.

The GAN is optimized by a minimax game, which minimizes the loss of $G$ but maximizes the loss of $D$.
We also note that the generator and discriminator are both incorporated with class embeddings during training.
This is a typical method of conditional GANs \cite{mirza2014conditional} that introduce external information to guide the training of GANs,
which is completely consistent with generative ZSL -- synthesizing visual features based on the side information of classes.
In addition, unlike most proposed generative methods that introduce auxiliary regularization or networks to ensure the inter-class discrimination of generated features, 
we implement our feature generation module with basic Wasserstein GAN and classification loss.
The core of our KG-GAN is to produce discriminative visual features conditioned on diverse and characteristic class semantics from KG.

\subsection{Softmax Classifiers for Unseen Classes}
At training stage,
the image features and class embeddings of seen classes are used to train the GAN model.
Once well trained, the KG-GAN is able to synthesize visual features of unseen classes from random noises with their corresponding class embeddings, since these unseen classes are semantically related to seen classes in knowledge graph.
Consequently, with synthesized unseen data $\hat{\mathcal{X}_u}$, we can learn a softmax classifier for each unseen class  and classify its testing samples.
The classifier is optimized by:
\begin{equation}
	\min \limits_\theta - \frac{1}{|\mathcal{X}|}  \sum_{(x,y) \in (\mathcal{X}, \mathcal{Y})} logP(y|x; \theta)
\end{equation}
where $\mathcal{X}$ represents the image features for training, $\mathcal{Y}$ is the label set to be predicted, $\theta$ is the training parameter and $P(y|x;\theta) = \frac{exp(\theta_y^Tx)}{\sum_i^{|\mathcal{Y}|} exp(\theta_i^Tx)}$.
Regarding the different prediction setting, $\mathcal{X}=\hat{\mathcal{X}_u}$ when it is ZSL and $\mathcal{X}=\mathcal{X}_s \cup \hat{\mathcal{X}_u}$ when it is GZSL, while the label set \cjy{$\mathcal{Y}$} is set to $\mathcal{Y}_u$ and $\mathcal{Y}_s \cup \mathcal{Y}_u$ respectively.

\section{Experiments}
We now perform experiments on image classification task to evaluate our proposed KG-GAN.
Firstly,
we compare KG-GAN against the state-of-the-art baselines in ZSL and GZSL setting,
and then we explore whether the rich semantics from our KG is more effective than other side information like textual descriptions (i.e., class word vectors).
We also analyze the impact of the {\it class embeddings} learned from different views of KG,
\cjy{and validate that incorporating semantic embeddings with the basic GAN is superior than incorporating additional regularization which is widely used in those baselines.}

\setlength{\tabcolsep}{4pt}

\renewcommand\arraystretch{1.1}
\begin{table}
\scriptsize

\begin{center}
\caption{Statistics of the two proposed datasets.}\label{tab:dataset}

\begin{tabular}{c|c|c|c|c|c}
    \hline 
    
	\bf {Dataset}
	& \bf{\tabincell{c}{Classes \\ \#} }
	& \bf{\tabincell{c}{Seen \\ Classes \#}}
	& \bf{\tabincell{c}{ Unseen \\ Classes \#}}        
	& \bf{\tabincell{c}{ Attributes  \\ \#} }
	& \bf{\tabincell{c}{ Total \\ Images \#}}
	\\ 
\hline
\tabincell{c}{ImNet-A}
    & 80  & 25  & 55 & 76  & 77,173
\\ \hline
\tabincell{c}{ImNet-O}
    & 35  & 10  & 25 & 38  & 39,361
\\    \hline

 \end{tabular}
\end{center}
\end{table}
\setlength{\tabcolsep}{1.4pt}

\subsection{Experiment Setting}
\subsubsection{Datasets} 
We extract evaluation data from widely used benchmark ImageNet \cite{DBLP:conf/cvpr/DengDSLL009}.
It is a large scale image classification dataset including a total of 21K classes and these classes are hierarchically related as the taxonomy structure stored in WordNet \cite{miller1995wordnet}.
Predicting on ImageNet is challenging since $1,000$ classes are taken as {\it seen} classes that have training samples but about 20K classes without training samples are taken as {\it unseen} classes.
Moreover, ImageNet contains a collection of fine-grained datasets as well as coarse-grained datasets.
The classes from fine-grained subset are usually grouped into different families, e.g. different vehicle types and different bird types.

To study the semantic relationship between classes in ZSL, 
we extract two datasets from the fine-grained subset for evaluation.
 One is domain-specific for animal classification (i.e., ImNet-A) and the other is general for object classification (i.e., ImNet-O).
The dataset split follows the original {\it seen-unseen} split proposed in \cite{DBLP:journals/pami/XianLSA19}.
Statistically, there are $25$ seen classes in ImNet-A, each of which contains around $1300$ images as training samples, and there are $55$ unseen classes which are ancestors, descendants or siblings of seen classes but without training samples.
Similarly, in ImNet-O, $10$ classes are taken as seen classes and $25$ as unseen classes.
The details of dataset are listed in Table~\ref{tab:dataset}.

\subsubsection{Knowledge Graph Construction}
We adopt the original taxonomy structure of WordNet as the backbone of knowledge graph, in which ImageNet classes are extracted and connected with each other via {\it subClass} relation, as
Fig.~\ref{fig:framework} shows.
Moreover, as attributes of ImageNet classes are not available, we invite volunteers to manually annotate attributes for these classes.
Specifically,  
for each class, annotators are asked to assign $3 \sim 6$ attributes from an attribute list\footnote{The list is collected from attribute annotations of well-known ZSL datasets, e.g., AWA, CUB and SUN, and appearance descriptions of classes from Wikipedia.} with $25$ images as references.
Each class is reviewed by $3 \sim 4$ volunteers, and we take consensus among the annotators as the final annotations.
Finally, we annotate $76$ attributes for ImNet-A classes and $38$ attributes for ImNet-O classes.
We add these attributes into knowledge graph, and link them with corresponding class nodes via {\it hasAttribute} relation.
The details of constructed KG are attached in the supplemental material.

\subsubsection{Baselines and Metrics}
We adopt both classic and the state-of-the-art ZSL methods as baselines.
Specifically, {\bf DeViSE} \cite{frome2013devise}, {\bf CONSE} \cite{norouzi2013zero} and {\bf SAE} \cite{DBLP:conf/cvpr/KodirovXG17} are  methods that map visual features into semantic space, and predict the labels of testing samples by computing nearest neighbor on class name word vectors;
Methods like {\bf SYNC} \cite{DBLP:conf/cvpr/ChangpinyoCGS16}, {\bf GCNZ} \cite{DBLP:conf/cvpr/0004YG18} and {\bf DGP} \cite{DBLP:conf/cvpr/KampffmeyerCLWZ19} propose to map class name word vectors into visual space to learn classifiers for unseen classes.
Note that GCNZ and DGP are two state-of-the-art ZSL methods that utilize KGs.
While {\bf GAZSL} \cite{DBLP:conf/cvpr/ZhuEL0E18}, {\bf LisGAN} \cite{DBLP:conf/cvpr/LiJLD0H19} and {\bf ZSL-ABP} \cite{zhu2019learning} are generative methods which generate visual features conditioned on class name word vectors (i.e., utilizing textual descriptions).
For fair comparisons, we re-evaluate these baselines on our proposed datasets and re-implement all methods in the same setting\footnote{The implementations of our method and all these baselines, as well as our data sets will be released if the paper is accepted.}.

Following previous work \cite{DBLP:journals/pami/XianLSA19}, we evaluate baselines and KG-GAN with {\bf Hit@k}, i.e., the ratio of samples whose top $k$ predicted labels hit the ground-truth label.
Considering the unbalanced number of samples of unseen classes in ImageNet, we compute the Hit@k independently for each class and average them as the results. 
In ZSL testing setting, the result is the average of Hit@k of each unseen class.
While in GZSL, we compute the harmonic mean $H = (2 * H_s * H_u)/(H_s + H_u)$, where $H_s$ and $H_u$ represent the average per-class Hit@k on seen classes and unseen classes respectively. 
Notably, we set $k$ to $1, 2, 5$ in ZSL setting, and set $k$ to $1$ in GZSL setting. 

\subsubsection{Implementation}
We employ well-performed CNN model ResNet101 \cite{DBLP:conf/cvpr/HeZRS16} to extract $2,048$-dimensional visual features for images, which is pre-trained on the samples of seen classes in ImageNet.
As for class name word vectors, we use pre-trained word embeddings provided by Changpinyo et al. \cite{DBLP:conf/cvpr/ChangpinyoCGS16}, they train skip-gram model on Wikipedia corpus and learn a $500$-dimensional word vector for each class.
Since there is no pre-trained attribute name word embeddings, we train them on Wikipedia corpus using Glove \cite{DBLP:conf/emnlp/PenningtonSM14} model.

In our KG-GAN, the encoder of GAE adopts a two-layer GCN.
After encoding the KG from two views, we obtain a $100$-dimensional {\it class embedding} for each class.
And we also set the dimension of noise vector $z$ to $100$. 
The generator and discriminator of GAN are both implemented with two fully connected layers, in which the generator has $4,096$ hidden units and outputs synthesized features with $2,048$ dimensions, and the discriminator also has $4,096$ hidden units and outputs a $2$-dimensional vector to indicate the input feature is real or not.
In training GAE module, the learning rate is set to $0.001$,
while in training GAN, the learning rate is set to $0.0001$.
We set the weight $\lambda$ for classification loss to $0.01$, and the weight $\beta$ for gradient penalty to $10$.


\setlength{\tabcolsep}{5pt}
\begin{table}
\scriptsize

\begin{center}

\caption{Performance ($\%$) of KG-GAN and baselines on ImNet-A and ImNet-O in ZSL setting.
$\S$ and $\dagger$ indicate generative and non-generative methods respectively.
The best and the second best results are marked in bold and underlined respectively.
The Hit@2 and Hit@5 of ZSL-ABP are omitted due to its KNN-based classifier during prediction.}
\label{tab:ZSL_results}

\begin{tabular}{c|l|ccc|ccc}

\hline
\multirow{2}{*}{ \ } &
\multirow{2}{*}{{ Methods}} 
& \multicolumn{3}{c|}{{ ImNet-A}} 
& \multicolumn{3}{c}{{ ImNet-O}} 
\\ \cline{3-8}
&& Hit@1 & Hit@2   & Hit@5    & Hit@1  & Hit@2   & Hit@5  \\\hline
\multirow{6}{*}{ $\dagger$ } &
DeViSE \cite{frome2013devise}
& 14.42 & 20.08 & 39.34 
		& 14.52 & 22.79 & 41.63   \\
& CONSE \cite{norouzi2013zero}
& 20.28 & 32.58 & 48.64 
		& 12.41 & 23.30 & {\bf 86.82}   \\
&SAE \cite{DBLP:conf/cvpr/KodirovXG17}
& 18.84 & 32.42 & 50.92 
		& 14.84 & 26.83 & 51.47    \\
&SYNC \cite{DBLP:conf/cvpr/ChangpinyoCGS16}
& 20.52 & 35.86 & 59.97 
		& 18.58 & 33.99 & 65.48    \\
&GCNZ \cite{DBLP:conf/cvpr/0004YG18}
& 31.62 & 60.19 & \underline {93.45} 
		& 30.05 & 50.19 & 84.50    \\
&DGP \cite{DBLP:conf/cvpr/KampffmeyerCLWZ19}  
	&\underline {33.07} 
	&\underline {60.34}
	&{\bf  93.49}
	&\underline {31.23}
	&\underline {50.32}
	&85.82  \\\hline
\multirow{4}{*}{ $\S$ } &
GAZSL \cite{DBLP:conf/cvpr/ZhuEL0E18}
& 20.57 & 35.82 & 60.55
		& 19.40 & 35.93 & 68.58       \\

&LisGAN \cite{DBLP:conf/cvpr/LiJLD0H19}
& 21.00 & 34.44   & 59.29     
		& 20.20 & 33.34   & 61.98 \\
&ZSL-ABP \cite{zhu2019learning}
& 22.05 & - & -
& 22.18 & - & -
\\\cline{2-8}
&KG-GAN 
	& {\bf  39.24} 
	& {\bf  63.66}  
	& 93.10
	& {\bf  34.65}
	& {\bf  55.15}
	& \underline{86.48}    
\\\hline
\end{tabular}
\end{center}
\end{table}
\setlength{\tabcolsep}{1.4pt}

\subsection{Comparison with Baselines}

\subsubsection{ZSL Setting} 
We first report the zero-shot learning results in Table~\ref{tab:ZSL_results}.
It can be seen that our method achieves the best results on Hit@1 and Hit@2 one two datasets, and also achieves state-of-the-art results on Hit@5.
In particular, on ImNet-A, the performance is improved by $6.17\%$ over the state-of-the-art DGP on Hit@1 and by $3.32\%$ on Hit@2.
On ImNet-O, the performance is improved by $3.42\%$ and $4.83\%$ on Hit@1 and Hit@2 respectively.
It is widely believed that the Hit@1 and Hit@2 are relatively more important \cite{DBLP:journals/pami/XianLSA19}, because these two metrics indicate that the models are capable of predicting the label to it right position more accurately.

From these results, we can observe that these generative methods perform better than most mapping-based ones except for GCNZ and DGP. 
These mapping-based methods process the prediction of unseen testing samples via the mutual mapping between semantic space and visual space, while those generative methods directly generate training samples for unseen classes conditioned on class semantics.
In mapping-based methods,
we note that the performance of methods whose mappings are from visual space to semantic space, e.g., DeViSE, CONSE and SAE, is much lower than those from semantic space to visual space, e.g., SYNC.
This indicates that the mapping from visual space to semantic space is less competitive due to the {\it hubness} problem.
As for GCNZ and DGP, they are KG-based methods which take the hierarchical relationship of classes as auxiliary semantics to assist the mapping from class name word vectors to visual features, and have superior performance than other baselines that only take class name word embeddings as class semantics.
We can conclude that the performance of ZSL can be significantly improved with the enrichment of class semantics, especially when we introduce knowledge graph that contains various class side information into generative ZSL model, the best results are achieved.


\setlength{\tabcolsep}{4pt}
\begin{table}
\scriptsize
\begin{center}
\caption{The Hit@1 (i.e., accuracy $\%$) results of generalized ZSL. 
$H$ is the harmonic mean of per-class accuracy on seen classes and unseen classes.
$\S$ and $\dagger$ indicate generative and non-generative methods respectively.
The best and the second best results are marked in bold and underlined respectively.
}
\label{tab:GZSL_results}

\begin{tabular}{c|l|ccc|ccc}
\hline
\multirow{2}{*}{ \ } &
\multirow{2}{*}{{Methods}} & \multicolumn{3}{c|}{{ImNet-A}} & \multicolumn{3}{c}{{ImNet-O}} 
\\ \cline{3-8}
&& $H_s$  & $H_u$  & {\bf $H$}  & $H_s$ & $H_u$ & {\bf $H$}   \\
\hline
\multirow{6}{*}{ $\dagger$ } &
DeViSE \cite{frome2013devise}
& 62.64 & 0.76  & 1.50 & 64.00 & 3.61  & 6.84         \\
&CONSE \cite{norouzi2013zero}
& 86.40 & 0.00  & 0.00 & 62.00 & 0.00 & 0.00         \\
&SAE \cite{DBLP:conf/cvpr/KodirovXG17}
& 86.48 & 0.00  & 0.00 
& {\bf  92.60} & 0.16 & 0.32    \\
&SYNC \cite{DBLP:conf/cvpr/ChangpinyoCGS16}
& {\bf 88.72} 
& 0.00  & 0.00  & 62.53  & 0.00 & 0.00        \\
&GCNZ \cite{DBLP:conf/cvpr/0004YG18}
& 49.12 & 15.85 & 23.96  & 44.60 & 14.48     & 21.87   \\
&DGP \cite{DBLP:conf/cvpr/KampffmeyerCLWZ19}
& 44.72 
    & \underline{17.92}
    & \underline{25.59}  
    & 47.40 
    & \underline{19.00}     
    & \underline{27.13}   \\\hline
\multirow{4}{*}{ $\S$ } &
GAZSL \cite{DBLP:conf/cvpr/ZhuEL0E18}
& \underline{86.56} & 1.28  & 2.52   
		& \underline{86.80} & 6.16 & 11.50   \\

&LisGAN \cite{DBLP:conf/cvpr/LiJLD0H19}
&  39.84   
	& 13.82  & 20.52     
	& 35.00  & 13.87  & 19.87 \\
&ZSL-ABP \cite{zhu2019learning}
& 53.60 & 12.11 & 19.75
& 49.20 & 12.81 & 20.33 
\\\cline{2-8}
&KG-GAN
	& 39.28
	& {\bf  25.06} 
	& {\bf  30.60} 
	& 43.40  
	& {\bf 20.82}   
	& {\bf  28.14}      
\\\hline
\end{tabular}
\end{center}
\end{table}
\setlength{\tabcolsep}{1.4pt}

\subsubsection{GZSL Setting}

We further evaluate the results of generalized ZSL in Table~\ref{tab:GZSL_results}, from which we can draw a similar conclusion as Table~\ref{tab:ZSL_results}.
Our KG-GAN performs better than listed methods and obtains significant outperformance on the prediction of unseen testing samples ($H_u$) and harmonic mean value ($H$), which means our KG-GAN has a better generalized capability.
However, we notice that the performance of mapping-based methods dramatically drop.
Methods like CONSE and SYNC even drop to $0.00$ on two datasets.
This illustrates that these methods have strong bias towards seen classes during prediction, i.e., the model tends to predict the unseen testing samples with labels from seen class set even if the label space contains unseen classes, which may be because these models overfit the training data of seen classes and can not generalize well on unseen classes.
By contrast, these generative methods weaken this trend and arise the performance on unseen classes.
We also find that although our framework does not achieve the best results on the prediction of seen testing samples ($H_s$), it still accomplishes comparable performance with the state-of-the-arts.
This motivates us to explore optimized algorithms to predict unseen testing samples correctly as well as maintain the high accuracy on seen classes.

\subsection{Class Semantics Analysis}

\begin{figure}
\scriptsize{
\centering
\subfigure[\scriptsize{ZSL}]{
    \begin{minipage}[b]{0.45\columnwidth}
    \includegraphics[width=1\columnwidth]{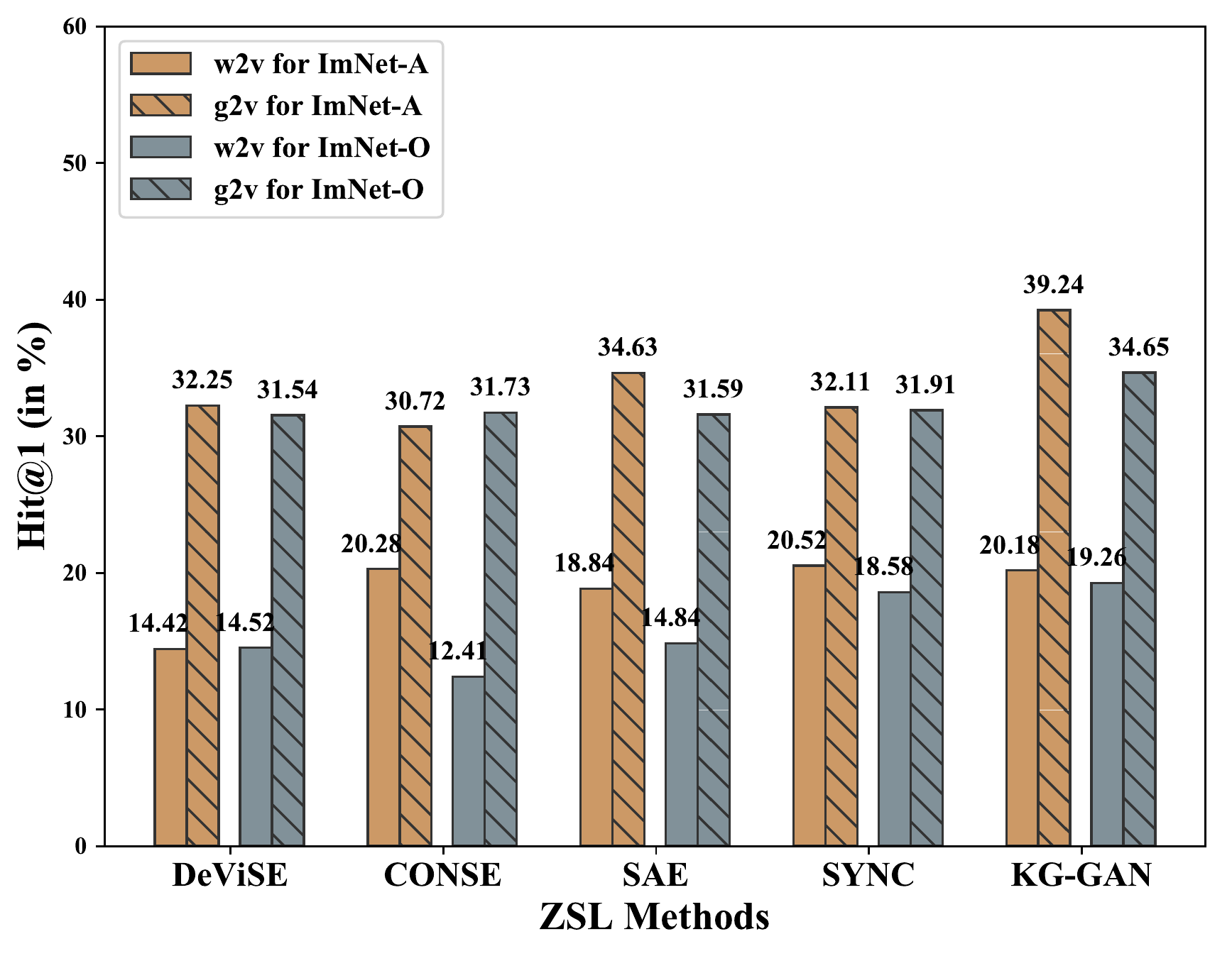}
    \end{minipage}
}
\subfigure[\scriptsize{Generalized ZSL}]{
    \begin{minipage}[b]{0.45\columnwidth}
    \includegraphics[width=1\columnwidth]{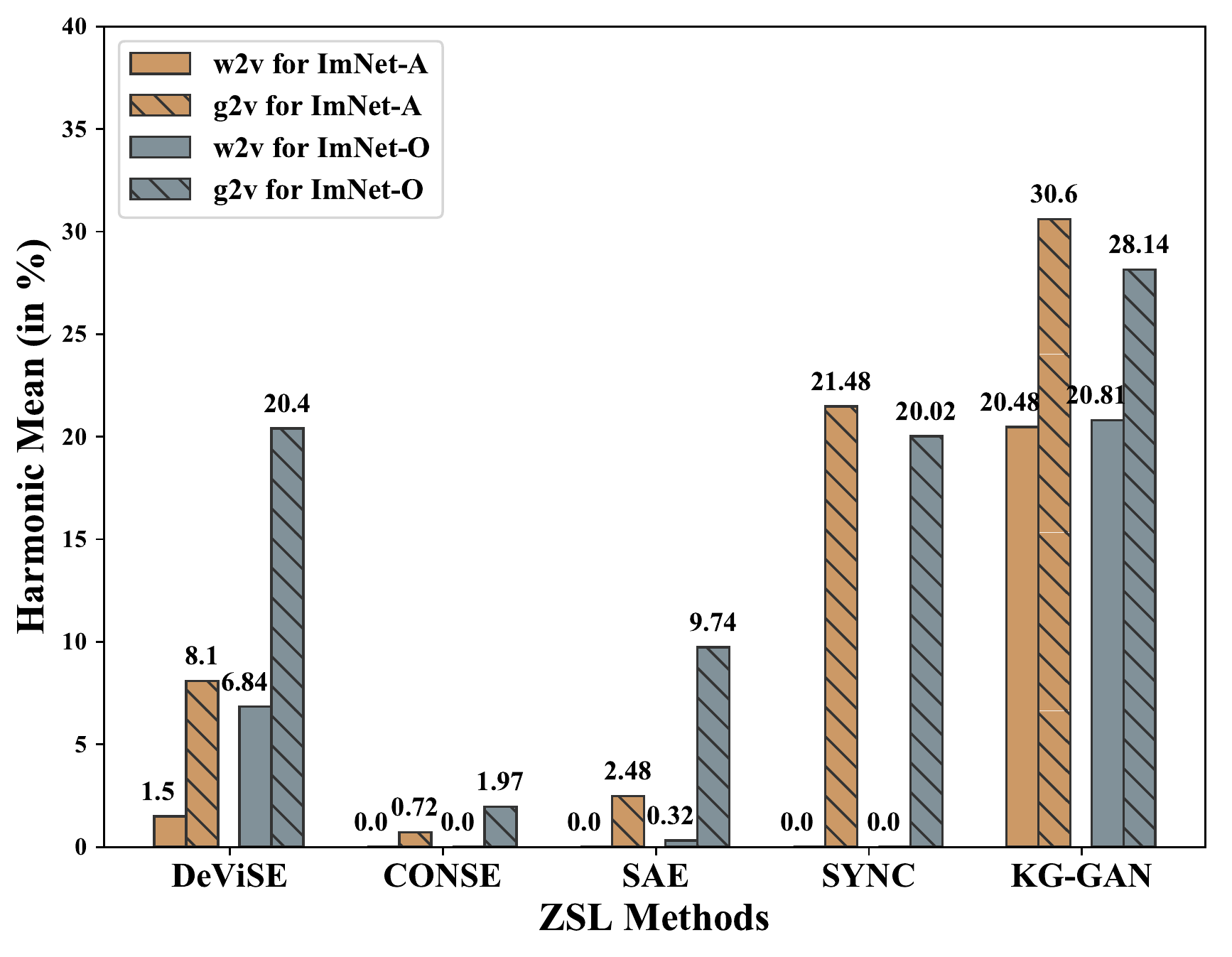}
    \end{minipage}
}
\caption{Performance comparison of mapping-based baselines and KG-GAN in different class semantics setting.
w2v: word2vec based class embedding; g2v: knowledge graph based class embedding.
In ZSL setting, we report the Hit@1 results.
} \label{fig:class_embed_results}
}
\end{figure}

To validate the superiority of our knowledge graph based class semantics, we replace the class embeddings of mapping-based methods (i.e., class name word vectors) with the class embeddings learned in our model (cf. Section \ref{class_embed}) and retrain these models to predict unseen testing samples.
In Fig.~\ref{fig:class_embed_results}, we report the comparison results of mapping-based baselines and our KG-GAN.
We find that the performance of all baselines has a significant improvement no matter what mapping direction the model is.
All of these methods have a more than $10\%$ increment on two datasets in ZSL setting, and more than $6\%$ in GZSL setting, especially for SYNC whose harmonic mean value increases from $0\%$ to $21.48\%$ on ImNet-A and from $0\%$ to $20.02\%$ on ImNet-O. 
On the other hand, our KG-GAN taking class name word embeddings as inputs expectedly performs worse due to the limited class semantics.
We also enhance the class semantics in previous KG-based methods (i.e., GCNZ and DGP).
Specifically, we add attribute nodes produced in our method into the KG they used to learn unseen classifiers.
As a result, taking DGP as an example, its performance is improved by $3.21\%$ on ImNet-A and by $2.72\%$ on ImNet-O respectively in ZSL setting, and improves $5.26\%$ on ImNet-A and $0.93\%$ on ImNet-O in GZSL setting.
To sum up, the knowledge graph which involves rich semantics about seen and unseen classes is of great advantage for ZSL problem, and class embeddings learned from it can remedy the weakness of ZSL models to some extent.

\subsection{Impact of KG Views}
In this subsection, we analyze the contribution of different views of KG for learning class embedding.
Specifically, we separately take class embeddings of different views, i.e., the {\it class}-view class embedding $g^c$ (denoted as GC) and {\it attribute}-view class embedding $g^a$ (denoted as GA), 
as the input of KG-GAN to synthesize visual features,
and evaluate the quality of generated features of different views.
We present the results in the last line of Table~\ref{tab:ablation},
from which we can see that the {\it attribute}-view class embedding has superior performance compared with the {\it class}-view one in ZSL and GZSL setting.
The higher performance may be due to
(i) the attribute annotations describe discriminative visual characteristics of class object, enabling different categories can be distinguished in ZSL model, especially for those having similar appearance;
(ii) our proposed datasets are fine-grained, which contains some sibling classes whose differences in taxonomy are not obvious such as {\it horse} and {\it zebra} in Fig.~\ref{fig:example}, making it difficult to differentiate the testing samples of these classes when only considering the {\it class}-view class embedding.
However, when combining these two class embeddings, we can achieve impressive prediction results, illustrating that the different views of class semantics from our knowledge graph is meaningful and complementary.

\setlength{\tabcolsep}{4pt}
\begin{table}
\scriptsize
\begin{center}
\caption{
The results of KG-GAN with different class embeddings learned from different views of KG, and the comparison results of generative methods with regularizer or not.
GC and GA represent the class embedding from {\it class}-view KG and {\it attribute}-view KG respectively, w2v is the class name vector originally used in these baselines, and reg refers to the model optimized with regularization.
}
\label{tab:ablation}
\begin{tabular}{lccc|cc}
\hline
\multirow{3}{*}{{ Methods}}   
& \multirow{3}{*}{{\tabincell{c}{ Setting}}}
& \multicolumn{2}{c|}{{ ZSL}} 
& \multicolumn{2}{c}{{ Generalized ZSL}} \\
&   & ImNet-A   & ImNet-O  & ImNet-A  & ImNet-O \\
&   & Hit@1      & Hit@1    & $H$   & $H$          \\\hline
 
 \multirow{3}{*}{GAZSL \cite{DBLP:conf/cvpr/ZhuEL0E18}}     
 & w2v + reg  & 20.57   & 19.40  & 2.52  & 11.50   \\
 & GC + reg   & 32.23   & 31.52  & 8.38 & 24.18          \\
 & GA + reg   & 36.32   & 33.43 & 26.82 & 24.87      \\\hline

 \multirow{3}{*}{LisGAN \cite{DBLP:conf/cvpr/LiJLD0H19}}

 & w2v + reg  & 21.00  & 20.20  & 20.52  & 19.87   \\
 & GC + reg   & 31.08  & 31.27  & 25.01  & 24.63   \\
 & GA + reg   & 36.96  & 33.45  & 29.43  & 25.75   \\\hline

\multirow{3}{*}{KG-GAN}    

& GC    & 32.95  & 32.18  & 26.06   &  26.50     \\
& GA    & 36.82  & 34.22  & 29.18   &  27.09          \\
& GC+GA   & {\bf 39.24}    & {\bf 34.65}   & {\bf 30.60}   & {\bf 28.14}         \\\hline

\end{tabular}
\end{center}
\end{table}
\setlength{\tabcolsep}{1.4pt}





\subsection{Regularizers or Class Semantics?}
In the literature of generative ZSL methods, some of them incorporate well-designed regularizers with GAN to improve the quality of synthesized features.
For example, in our baselines, GAZSL exploits visual pivot to regularize the mean of generated features of each class to be the mean of real feature distribution; 
LisGAN regularizes the generated samples to be close to the soul samples, which represents the multi-view prototype representation of real samples.
In contrast,
our KG-GAN only relies on rich class semantics to synthesize discriminative features for classes without any regularization.
To compare the effectiveness of regularizers and class semantics in generating high-quality samples, we input two class embeddings learned from two-views KG (i.e., GC and GA) into the above generative models, and analyze the prediction results in ZSL and GZSL setting.
As Table~\ref{tab:ablation} shows,
our method combining GC and GA outperforms the baselines combining GC (GA) and regularizers.
This indicates that rich class semantics may be more important than optimized regularizers for generative ZSL model.

 





\section{Conclusions}
In this paper, we propose to leverage a knowledge graph with two views as class semantics to synthesize sample features in generative zero-shot learning using Generative Adversarial Networks.
Our method KG-GAN
achieves promising results on different datasets from ImageNet, outperforming the state-of-the-art in both none generative ZSL and generative ZSL.
It uses Graph Auto-Encoder to learn semantically meaningful class embeddings,
which are superior than traditional class name word vectors and class hierarchical relationship in semantics,
and are superior than well-designed regularization in generating discriminative visual features.
This inspires us to explore more potential side information and build richer knowledge graphs to tackle the problem of ZSL, especially for improving the performance on both seen and unseen classes.

%
%
\bibliographystyle{splncs04}
\bibliography{eccv20}
\end{document}